\documentclass[11pt]{article}
\usepackage{acl2010}
\usepackage{url}
\usepackage{times}
\usepackage{latexsym}
\usepackage{xspace}
\usepackage{url}
\usepackage{color}
\usepackage[small,compact]{titlesec}
\usepackage{multirow}
\usepackage{microtype}
\newcommand{\other}{\textsc{Other}\xspace}
\newcommand{\secref}[2][]{Section#1~\ref{#2}\xspace}
\newcommand{\tabref}[2][]{Table#1~\ref{#2}\xspace}

\title{SemEval-2010 Task 8: Multi-Way Classification \\ of Semantic Relations Between Pairs of Nominals}

\author{
Iris Hendrickx\thanks{\hspace{.1cm}  University of Lisbon, \scriptsize{iris@clul.ul.pt}}\hspace{.2cm},
Su Nam Kim\thanks{\hspace{.1cm}  University of Melbourne, \scriptsize{snkim@csse.unimelb.edu.au}}\hspace{.2cm},
Zornitsa Kozareva\thanks{\hspace{.1cm} University of Alicante, \scriptsize{zkozareva@dlsi.ua.es}}\hspace{.2cm},
Preslav Nakov\thanks{\hspace{.1cm} National University of Singapore, \scriptsize{nakov@comp.nus.edu.sg}}\hspace{.2cm},\\
\textbf{Diarmuid \'O S\'eaghdha\thanks{\hspace{.1cm} University of Cambridge, \scriptsize{do242@cl.cam.ac.uk}}\hspace{.2cm},
Sebastian Pad\'o\thanks{\hspace{.1cm} University of Stuttgart, \scriptsize{pado@ims.uni-stuttgart.de}}\hspace{.2cm},
Marco Pennacchiotti\thanks{\hspace{.1cm} Yahoo! Inc., \scriptsize{pennacc@yahoo-inc.com}}\hspace{.2cm},}\\
\textbf{Lorenza Romano\thanks{\hspace{.1cm}\hspace{.1cm} Fondazione Bruno Kessler, \scriptsize{romano@fbk.eu}}\hspace{.2cm},
Stan Szpakowicz\thanks{\hspace{.1cm}\hspace{.1cm} University of Ottawa \emph{and} Polish Academy of Sciences, \scriptsize{szpak@site.uottawa.ca}}}\hspace{.2cm}
}

\begin{document}
\maketitle

\begin{abstract}

In response to the continuing research interest in computational semantic analysis,
we have proposed a new task for SemEval-2010:
multi-way classification of mutually exclusive
semantic relations between pairs of nominals.
The task is designed to compare different approaches to the problem and to
provide a standard testbed for future research.
In this paper, we define the task, describe the creation of the
datasets, and discuss the results of the participating 28 systems submitted by
10 teams.
\end{abstract}

\section{Introduction}
\label{sec:intro}

SemEval-2010 Task 8 focuses on \textit{semantic relations between
  pairs of nominals}. For example,
  \emph{tea} and \emph{ginseng} are in an
\textsc{Entity-Origin} relation in ``\emph{The cup contained
\underline{tea} from dried \underline{ginseng}.}''.
The automatic recognition of semantic relations can have many
applications, such as information extraction, document
summarization, machine translation, or construction of thesauri and
semantic networks. It can also facilitate auxiliary tasks such as word
sense disambiguation, language modeling, paraphrasing, and recognizing
textual entailment.

Our goal in SemEval-2010 Task 8 was to create a testbed for automatic classification
of semantic relations. In developing the task we met
several challenges: selecting a suitable set of relations, specifying
the annotation procedure, and deciding on the details of the task
itself. They are discussed briefly in \secref{sec:data}; see also
Hendrickx et al.~\shortcite{hendrickx09:semeval_2010_task_8}, which
includes a survey of related work. The direct predecessor of Task 8 was
\textit{Classification of semantic relations between nominals}, Task 4
at SemEval-1
\cite{girju09:_class_of_seman_relat_between_nomin}. That task
 had a separate binary-labeled dataset for each of seven
relations. We have defined SemEval-2010 Task 8 as multi-way
classification, where the label for each example must be chosen
from the complete set of ten relations. We have also produced
a larger quantity of data: 10,717 annotated examples, contrasted with
1,529 in SemEval-1 Task 4.

\section{Dataset Creation}\label{sec:data}

\subsection{The Inventory of Semantic Relations}

We first decided on an inventory of semantic relations. Ideally, it
should be exhaustive (enable the description of relations between any
pair of nominals) and mutually exclusive (each pair of nominals
\emph{in context} should map onto only one relation). The literature,
however, suggests that no relation inventory satisfies both needs,
and, in practice, some trade-off between them must be accepted.

As a pragmatic compromise,
we selected nine relations with coverage sufficiently broad to be of
general and practical interest. We aimed at avoiding semantic overlap
as much as possible. We included, however, two groups of strongly
related relations (\textsc{En\-ti\-ty-Or\-i\-gin} /
\-\textsc{En\-ti\-ty-\-Des\-ti\-na\-tion} and
\textsc{Con\-tent-Con\-tain\-er }/ \textsc{Com\-po\-nent-Whole} /
\-\textsc{Member-Collection}), which can help assess models'
ability to make such fine-grained distinctions. Our inventory is given
below. The first four relations appeared in SemEval-1 Task 4, but the
annotation guidelines have been revised, and thus no complete
continuity should be assumed.

\begin{description}
\item[Cause-Effect (CE).] An event or object yields an effect. Example:
  \textsl{those cancers were caused by radiation exposures}
\item[Instrument-Agency (IA).] An agent uses an instrument. Example:
  \textsl{phone operator}
\item[Product-Producer (PP).] A producer causes a product to
  exist. Example: \textsl{a factory manufactures suits}
\item[Content-Container (CC).] An object is physically stored in a delineated area of space. Example: \textsl{a bottle full of honey was weighed}
\item[Entity-Origin (EO).] An entity is coming or is derived from an origin (e.g., position or material). Example: \textsl{letters from foreign countries}
\item[Entity-Destination (ED).] An entity is moving towards a destination.
     Eg. \textsl{the boy went to bed}
\item[Component-Whole (CW).] An object is a component of a larger
 whole. Example: \textsl{my apartment has a large kitchen}
\item[Member-Collection (MC).] A member forms a nonfunctional part of a collection.
Example: \textsl{there are many trees in the forest}
\item[Communication-Topic (CT).] An act of communication, written
  or spoken, is about a topic. Example: \textsl{the lecture was about
    semantics}
\end{description}

\subsection{Annotation Guidelines}

We defined a set of general annotation guidelines
as well as detailed guidelines for each semantic relation.
Here, we describe the general guidelines, which delineate the
scope of the data to be collected and state general principles
relevant to the annotation of all relations.\footnote{The full task
  guidelines are available at
  \url{docs.google.com/View?id=dfhkmm46\_0f63mfvf7}}

Our objective is to annotate instances of semantic relations which are
true in the sense of holding in the most plausible truth-conditional
interpretation of the sentence. This is in the tradition of the
Textual Entailment or Information Validation paradigm~\cite{RTE}, and
in contrast to ``aboutness'' annotation such as semantic
roles~\cite{Carreras:Marquez:04,Carreras:Marquez:05}
or the BioNLP 2009 task \cite{Kim:EtAl:09}
where negated relations are also labelled as positive. Similarly, we
exclude instances of semantic relations which hold only in hypothetical
or counterfactural scenarios. In practice, this means disallowing
annotations within the scope of modals or negations,
e.g., ``\emph{Smoking \underline{may/may not} have caused cancer in this case.}''

We accept as relation arguments only noun phrases with common-noun heads.
This distinguishes our task from much work in
Information Extraction, which tends to focus on specific classes of
named entities and on considerably more fine-grained relations than we
do.  Named entities are a specific category of nominal expressions
best dealt with using techniques which do not apply to common nouns.
We only mark up the semantic heads of nominals, which usually span a single
word, except for lexicalized terms such as \textit{science fiction}.

We also impose a syntactic locality requirement on example candidates,
thus excluding instances where the relation arguments occur
in separate sentential clauses. Permissible syntactic patterns include
simple and relative clauses, compounds, and pre- and post-nominal
modification. In addition, we did not annotate examples whose
interpretation relied on discourse knowledge, which led to the
exclusion of pronouns as arguments. See the
guidelines for details on other issues (noun
compounds, aspectual phenomena, temporal relations).

\subsection{The Annotation Process}

The annotation took place in three rounds.  First, we manually
collected for each relation
around
1,200 sentences
through pattern-based Web search.  In order to ensure a wide variety of
example sentences, we used a substantial number of patterns for each
relation, typically between one hundred and several hundred.
Importantly, in the first round, the relation itself was not annotated:
the goal was merely to collect positive and near-miss candidate
instances. A rough aim was to have 90\% of candidates which
instantiate the target relation (``positive instances'').

In the second round, the collected candidates for each relation went
to two independent annotators for labeling.
Since we have a multi-way classification task, the annotators used the full inventory of nine relations plus
\other. The annotation was made easier by the fact that
the cases of overlap were largely systematic, arising from general
phenomena like metaphorical use and situations where more than one relation holds.
 For example, there is a systematic potential overlap between
\textsc{Content-Container} and \textsc{Entity-Destination} depending
on whether the situation described in the sentence is static or dynamic,
e.g., ``\emph{When I came, the $<$e1$>$apples$<$/e1$>$ were already put in the
$<$e2$>$basket$<$/e2$>$.}'' is \textsc{CC}(e1, e2),
while ``\emph{Then, the $<$e1$>$apples$<$/e1$>$ were quickly put in the $<$e2$>$basket$<$/e2$>$.}''
is \textsc{ED}(e1, e2).

In the third round, the remaining disagreements were resolved, and, if
no consensus could be achieved, examples were removed. Finally, we
collected all examples for the individual relation datasets and merged
them into a final dataset of 10,717 instances: we released 8,000 of
them for training and kept the remainder for testing.\footnote{This
  set includes 891 examples from SemEval-1 Task 4.  We re-annotated
  them and assigned them as the last examples of our \emph{training}
  dataset to ensure that the test set was unseen.}

\tabref{tab:anno-stats} shows more statistics about the
dataset. The second row (Pos) shows that the average share of positive
instances was closer to 75\% than to 90\%, indicating that the
patterns catch a substantial amount of ``near-miss'' cases.
This effect, however, varies a lot across relations, causing the non-uniform
relation distribution in the test set (first row). After the second
round, we also computed inter-annotator agreement (third column, IAA)
at the sentence level as the percentage of sentences for which the two
annotations were identical. We do not report Kappa, since chance
agreement on our preselected candidate datasets is difficult to
estimate. IAA is between 60\% and 95\%, again with large
relation-dependent variation.  Some of the relations were particularly
easy to annotate, notably \textsc{Content-Container}, despite the
systematic ambiguity mentioned above, which can be resolved through
relatively clear criteria. \textsc{Entity-Origin} was the hardest
relation to annotate. We encountered ontological difficulties in
defining both Entity (e.g.,\ in contrast to Effect) and Origin (as
opposed to Cause). Our numbers are on average around 10\% higher than
those reported by Girju et
al.~\shortcite{girju09:_class_of_seman_relat_between_nomin}. This may
be a side effect of our data collection method. To gather 1,200
examples in realistic time, we had to seek productive search query
patterns, which invited certain homogeneity. For example, many queries
for \textsc{Content-Container} centered on ``usual suspect'' such as
\emph{box} or \emph{suitcase}. Many instances of
\textsc{Member-Collection} arose from available lists of collective names.

\begin{table}
  \centering
  \small
  \begin{tabular}{l|ccc}
    Relation & Freq & Pos & IAA  \\ \hline
    Cause-Effect &  1331 (12.4\%)  & 91.2\%  &  79.0\%  \\ 
    Component-Whole &  1253 (11.7\%) & 84.3\% & 70.0\%  \\
    Entity-Destination &  1137 (10.6\%)& 80.1\% & 75.2\% \\
    Entity-Origin &   974 (9.1\%) & 69.2\% &    58.2\%  \\
    Product-Producer &   948 (8.8\%) & 66.3\% & 84.8\%  \\
    Member-Collection &   923 (8.6\%) & 74.7\% &68.2\%  \\
    Message-Topic &   895 (8.4\%) & 74.4\% &    72.4\%  \\
    Content-Container &   732 (6.8\%) & 59.3\% &95.8\%  \\
    Instrument-Agency &   660 (6.2\%) & 60.8\% &65.0\%  \\
    Other &  1864 (17.4\%)  \\ \hline \hline
    Total & 10717 (100\%)
  \end{tabular}
  \caption{Annotation Statistics. Freq: Absolute and relative frequency in the dataset; Pos: percentage of ``positive'' relation instances in the candidate set; IAA: inter-annotator agreement}
  \label{tab:anno-stats}
\end{table}

\section{The Task}
\label{sec:task}

The participating systems had the following task:
given a sentence and two tagged nominals,
predict the relation between those nominals \emph{and} the direction of the relation.

We released a detailed scorer which outputs (1)~a confusion matrix, (2)
accuracy and coverage, (3) precision (P), recall (R), and F1-score for
each relation, (4) micro-averaged P, R, F1, (5) macro-averaged P, R,
F1.  For (4) and (5), the calculations ignored the \other relation.
Our official scoring metric is macro-averaged F1-score for (9+1)-way
classification, taking directionality into account.

The teams were asked to submit test data predictions for varying fractions
of the training data. Specifically, we requested results for the first
1000, 2000, 4000, and 8000 training instances, called TD1 through
TD4 (TD4 was the full training set).

\section{Participants and Results}\label{sec:results}

\tabref{tab:participants} lists the participants and provides a
rough overview of the system features. \tabref{tab:results} shows
the results. Unless noted otherwise, all quoted numbers are F-scores.

\begin{table*}
  \centering
  \small
  \begin{tabular}{p{.18\textwidth}|p{.15\textwidth}p{.15\textwidth}|p{.25\textwidth}p{.05\textwidth}p{.07\textwidth}}
    System & Institution & Team  & Description & Res. & Class. \\ \hline\hline
     Baseline & Task organizers &   & local context of 2 words only &  &  BN\\ \hline
    ECNU-SR-1 &  East China Normal University & \multirow{3}{.15\textwidth}{Man Lan, Yuan Chen, Zhimin Zhou, Yu Xu} & stem, POS, syntactic patterns & S & SVM (multi) \\
    &&&\\ \hline
    ECNU-SR-2,3 & & & features like ECNU-SR-1, different prob. thresholds & & SVM (binary)    \\ \hline
    ECNU-SR-4 & & & stem, POS, syntactic patterns, hyponymy and meronymy relations & WN, S & SVM (multi) \\ \hline
    ECNU-SR-5,6 & & & features like ECNU-SR-4, different prob. thresholds  & & SVM (binary)  \\ \hline
    ECNU-SR-7 & & & majority vote of ECNU-1,2,4,5 \\ \hline\hline

    FBK\_IRST-6C32 & Fondazione Bruno Kessler & Claudio Giuliano, Kateryna Tymoshenko & 3-word window context features (word form, part of speech, orthography) + Cyc; parameter estimation by optimization on training set & Cyc & SVM \\ \hline
    FBK\_IRST-12C32 & & & FBK\_IRST-6C32 + distance features \\ \hline
    FBK\_IRST-12VBC32 & & & FBK\_IRST-12C32 + verbs  \\     \hline
    FBK\_IRST-6CA, -12CA, -12VBCA & & & features as above, parameter estimation by cross-validation \\ \hline \hline

    FBK\_NK-RES1 & Fondazione Bruno Kessler & Matteo Negri, Milen Kouylekov & collocations, glosses, semantic relations of nominals + context features & WN & BN \\ \hline
    FBK\_NK-RES 2,3,4 & &  & like FBK\_NK-RES1 with different context windows and collocation cutoffs \\  \hline\hline

    ISI & Information Sciences Institute, University of Southern California & Stephen Tratz & features from different resources, a noun compound relation system, and various feature related to capitalization, affixes, closed-class words & WN, RT, G & ME \\ \hline \hline

    ISTI-1,2 & Istituto di scienca e tecnologie dell'informazione ``A. Faedo'' &  Andrea Esuli, Diego Marcheggiani, Fabrizio Sebastiani &
    Boosting-based classification. Runs differ in their initialization. & WN & 2S \\ \hline \hline

    JU &  Jadavpur University & Santanu Pal, Partha Pakray, Dipankar Das, Sivaji Bandyopadhyay & Verbs, nouns, and prepositions; seed lists for semantic relations; parse features and NEs & WN, S & CRF \\ \hline \hline

    SEKA & Hungarian Academy of Sciences  & Eszter Simon, Andras Kornai & Levin and Roget classes, n-grams; other grammatical and formal features & RT, LC & ME \\ \hline \hline

    TUD-base & Technische Universit\"at Darmstadt & Gy\"orgy Szarvas, Iryna Gurevych &  word, POS n-grams, dependency path, distance & S & ME \\ \hline
    TUD-wp & & & TUD-base + ESA semantic relatedness scores & +WP \\  \hline
    TUD-comb & & & TUD-base + own semantic relatedness scores & +WP,WN \\\hline
    TUD-comb-threshold & & & TUD-comb with higher threshold for \other \\ \hline \hline

    UNITN & University of Trento & Fabio Celli &  punctuation, context words, prepositional patterns, estimation of semantic relation & -- & DR  \\ \hline \hline

    UTD & University of Texas at Dallas & Bryan Rink, Sanda Harabagiu & context wods, hypernyms, POS, dependencies, distance, semantic roles, Levin classes, paraphrases & WN, S, G, PB/NB, LC & SVM, 2S \\ \hline \hline

  \end{tabular}
  \caption{Participants of SemEval-2010 Task 8. Res: Resources used (WN: WordNet data; WP: Wikipedia data; S: syntax; LC: Levin classes; G: Google n-grams, RT: Roget's Thesaurus, PB/NB: PropBank/NomBank). Class: Classification style (ME: Maximum Entropy; BN: Bayes Net; DR: Decision Rules; CRF: Conditional Random Fields; 2S: two-step classification) }
  \label{tab:participants}
\end{table*}

\begin{table*}[tb]
  \centering
  \small
  \begin{tabular}{l|cccc|c|c|ll}
    System                  & TD1 & TD2 & TD3 & TD4 & Acc TD4 & Rank & Best Cat & Worst Cat-9    \\ \hline
    Baseline                   &33.04 & 42.41 & 50.89 & 57.52 & 50.0 & - & MC (75.1) & IA (28.0)\\ \hline
    ECNU-SR-1             & 52.13 & 56.58 & 58.16 & 60.08 & 57.1 & \multirow{7}{.02\textwidth}{4} & CE (79.7) & IA (32.2) \\
    ECNU-SR-2             & 46.24 & 47.99 & 69.83 & 72.59 & 67.1 & & CE (84.4) & IA (52.2) \\
    ECNU-SR-3             & 39.89 & 42.29 & 65.47 & 68.50 & 62.0 & & CE (83.4) & IA (46.5) \\
    ECNU-SR-4             & 67.95 & 70.58 & 72.99 & 74.82 & 70.5 & & CE (84.6) & IA (61.4)  \\
    \textit{ECNU-SR-5}    & 49.32 & 50.70 & 72.63 & 75.43 & 70.2 & & CE (85.1) & IA (60.7) \\
    ECNU-SR-6             & 42.88 & 45.54 & 68.87 & 72.19 & 65.8 & & CE (85.2) & IA (56.7) \\
    ECNU-SR-7             & 58.67 & 58.87 & 72.79 & 75.21 & 70.2 & & CE (86.1) & IA (61.8)  \\ \hline
    FBK\_IRST-6C32           & 60.19 & 67.31 & 71.78 & 76.81 & 72.4 & \multirow{6}{.02\textwidth}{2} & ED (82.6) & IA (69.4) \\
    FBK\_IRST-12C32          & 60.66 & 67.91 & 72.04 & 76.91 & 72.4 & & MC (84.2) & IA (68.8)\\
    FBK\_IRST-12VBC32        & 62.64 & 69.86 & 73.19 & 77.11 & 72.3 & & ED (85.9) & PP (68.1) \\
    FBK\_IRST-6CA            & 60.58 & 67.14 & 71.63 & 76.28 & 71.4 & & CE (82.3) & IA (67.7)\\
    FBK\_IRST-12CA           & 61.33 & 67.80 & 71.65 & 76.39 & 71.4 & & ED (81.8) & IA (67.5)\\
    \textit{FBK\_IRST-12VBCA}& 63.61 & 70.20 & 73.40 & 77.62 & 72.8 & & ED (86.5) & IA (67.3)  \\ \hline
    \textit{FBK\_NK-RES1} & 55.71$^*$ & 64.06$^*$ & 67.80$^*$ & 68.02 & 62.1 & \multirow{4}{.02\textwidth}{7} & ED (77.6) & IA (52.9) \\
    FBK\_NK-RES2          & 54.27$^*$ & 63.68$^*$ & 67.08$^*$ & 67.48 & 61.4 & & ED (77.4) & PP (55.2) \\
    FBK\_NK-RES3          & 54.25$^*$ & 62.73$^*$ & 66.11$^*$ & 66.90 & 60.5 & & MC (76.7) & IA (56.3) \\
    FBK\_NK-RES4          & 44.11$^*$ & 58.85$^*$ & 63.06$^*$ & 65.84 & 59.4 & & MC (76.1) & IA/PP (58.0)\\ \hline
    \textit{ISI}          & 66.68 & 71.01 & 75.51 & 77.57 & 72.7            & \multirow{1}{.02\textwidth}{3} & CE (87.6) & IA (61.5) \\ \hline
    \textit{ISTI-1}       & 50.49$^*$ & 55.80$^*$ & 61.14$^*$ & 68.42 & 63.2 & \multirow{2}{.02\textwidth}{6} & ED (80.7) & PP (53.8) \\
    ISTI-2                & 50.69$^*$ & 54.29$^*$ & 59.77$^*$ & 66.65 & 61.5 & & ED (80.2) & IA (48.9) \\ \hline
   \textit{JU}            & 41.62$^*$ & 44.98$^*$ & 47.81$^*$ & 52.16 & 50.2 &  \multirow{1}{.02\textwidth}{9} & CE (75.6) & IA (27.8) \\ \hline
    \textit{SEKA}         & 51.81 & 56.34 & 61.10 & 66.33 & 61.9 & \multirow{1}{.02\textwidth}{8} & CE (84.0) & PP (43.7) \\ \hline
    TUD-base              & 50.81 & 54.61 & 56.98 & 60.50 & 56.1 & \multirow{4}{.02\textwidth}{5} & CE (80.7) & IA (31.1) \\
    TUD-wp                & 55.34 & 60.90 & 63.78 & 68.00 & 63.5 & & ED (82.9) & IA (44.1) \\
    TUD-comb              & 57.84 & 62.52 & 66.41 & 68.88 & 64.6 & & CE (83.8) & IA (46.8)\\
    \textit{TUD-comb-$\theta$}& 58.35 & 62.45 & 66.86 & 69.23 & 65.4 & & CE (83.4) & IA (46.9) \\ \hline
    \textit{UNITN}        & 16.57$^*$ & 18.56$^*$ & 22.45$^*$ & 26.67 & 27.4 & \multirow{1}{.02\textwidth}{10} & ED (46.4) & PP (0) \\ \hline
    \textit{UTD}          & \textbf{73.08} & \textbf{77.02} & \textbf{79.93} & \textbf{82.19} & 77.9 & 1 & CE (89.6) & IA (68.5) \\ \hline
  \end{tabular}
  \caption{Performance (F$_1$ Score) of all submitted systems on the test dataset as a function of the size of the training dataset: TD1=1000, TD2=2000, TD3=4000, TD4=8000 training examples. The official results are calculated on TD4. The results marked with $^*$ were submitted after the deadline. The best-performing run for each participant is \emph{italicized}.}
  \label{tab:results}
\end{table*}

\paragraph{Overall Ranking and Training Data.} We ranked the teams by the
performance of their best system on TD4, since a per-system ranking
would favor teams with many submitted runs. UTD submitted the best
system, with a performance of over 82\%, more than 4\% better than the
second-best system. FBK\_IRST was placed second, with 77.62\%, a tiny margin
ahead of ISI (77.57\%). Notably, the ISI system outperforms the FBK\_IRST
system for TD1 to TD3, where it was second-best.
The accuracy numbers for TD4 (Acc TD4) lead to the same overall ranking:
micro- versus macro-averaging does not appear to make much
difference either. A random baseline gives an uninteresting score of 6\%.
Our baseline system is a simple Naive Bayes classifier which relies
on words in the sentential context only; two systems scored below this
baseline.

As for the amount of training data, we see a substantial improvement
for all systems between TD1 and TD4, with diminishing returns for the
transition between TD3 and TD4 for many, but not all,
systems. Overall, the differences between systems are smaller for TD4
than they are for TD1. The spread between the top three systems is around
10\% at TD1, but below 5\% at TD4. Still, there are clear
differences in the influence of training data size even among systems
with the same overall architecture. Notably, ECNU-SR-4 is the
second-best system at TD1 (67.95\%), but gains only 7\% from
the eightfold increase of the size of the training data. At the same time, ECNU-SR-3
improves from less than 40\% to almost 69\%. The difference between
the systems is that ECNU-SR-4 uses a multi-way classifier including
the class \other, while ECNU-SR-3 uses binary classifiers and
assigns \other if no other relation was assigned with $p$$>$0.5. It
appears that these probability estimates for classes are only reliable
enough for TD3 and TD4.

\paragraph{The Influence of System Architecture.} We investigate
the classification scheme and the resources
used.  Almost all systems used either MaxEnt or SVM classifiers,
with no clear advantage for either. Similarly, two systems, UTD and
ISTI (rank 1 and 6) split the task into two classification steps
(relation and direction), but the 2nd- and 3rd-ranked systems do
not. The use of a sequence model did not show a benefit either.

The systems use a variety of resources. Generally, richer feature sets
lead to better performance (although the differences are often small
-- compare the different FBK\_IRST systems). This can be explained by
the need for semantic generalization from training to test data. This
need can be addressed using WordNet (contrast ECNU-1 to -3 with ECNU-4 to
-6), the Google $n$-gram collection (see ISI and UTD), or a ``deep''
semantic resource (FBK\_IRST uses Cyc). Yet, most of these resources
are also included in the less successful systems, so it does not seem
easy to achieve a beneficial integration of knowledge sources in
systems for semantic relation classification.

\paragraph{System Combination.} The differences between the systems
suggest that it might be possible to achieve improvements by building an ensemble system.
Exploring this idea, we first combined the top three
systems (UTD, FBK\_IRST-12VBCA, and ISI) by outputting the majority vote,
or \other if there was none. This combination yielded small
improvement over the UTD system with an F-score of 82.79\%. For a
combination of the top five systems using the same method, the performance
decreased again, yielding 80.42\%. This suggests
that the best system outperforms the rest by a margin that
cannot be compensated with system combination, at least not with a
crude majority vote. We see a similar pattern among
the ECNU systems, where the ECNU-SR-7 combination system is
outperformed by ECNU-SR-5, presumably since it incorporates the rather
inferior ECNU-SR-1 system.

\paragraph{Relation-specific Analysis.} We also analyzed the
performance on individual relations, especially the extremes.
There are very stable patterns across all systems. The best
relation (presumably the easiest to classify) was CE, far ahead of ED and
MC. Notably, the performance for the best relation was 75\% or above for
almost all systems, with comparatively small differences between
the systems. The hardest relation was generally IA,
followed by PP.\footnote{The relation \other, which we ignore in
  the overall F-score, did even worse, often below 40\%. This is to
  be expected, since the \other examples in our datasets are near
  misses for other relations, thus making a very incoherent
  class.} Here, the spread among the systems is much larger, so perhaps
the highest-ranking systems outperform others on the
difficult relations. Recall was the main problem for both IA and PP:
 many examples of these two relations are misclassified,
most frequently as \other. Even at TD4, these
datasets seem to be less homogeneous than the others. Intriguingly,
PP shows a very high inter-annotator agreement
(\tabref{tab:anno-stats}). Its difficulty may therefore be
due not to questionable annotation, but either to genuine intrinsic
variability, or at least more varied pattern selection by the
dataset creator. Conversely, MC -- among the easiest relations to model --
had a relatively modest IAA.

\paragraph{Difficult Instances.} There were 152 examples that were classified incorrectly
by all systems. We analyzed them, looking for sources
of errors. In addition to a handful of annotation errors and some
borderline cases, they are made up of instances which illustrate the
limits of current shallow modeling approaches in that they require
more lexical knowledge and complex reasoning.  A case in point:
\textit{The bottle carrier converts your $<$e1$>$bottle$<$/e1$>$ into
  a $<$e2$>$canteen$<$/e2$>$}. This instance of \other has been
misclassified either as CC (due to the nominals) or as ED (because of
the preposition \emph{into}). Another example: \textit{Flanking or
  backing $<$e1$>$rudders$<$/e1$>$ are used by
  $<$e2$>$towboats$<$/e2$>$ and other vessels that require a high
  degree of manoeuvrability.} This is an instance of CW misclassified
as IA, probably on account of the verb \emph{use} which is a frequent
indicator of an agentive relation.

\section{Discussion and Conclusion}

There is little doubt that 19-way classification is a non-trivial
challenge. It is even harder when the domain is lexical semantics,
with its inherent idiosyncrasies, and the classes are not necessarily
disjoint, despite our best intentions. It speaks to the success of the
exercise that the participating systems' performance was generally
quite high, well over an order of magnitude above random guessing.
This may be due to the impressive
array of tools and lexical-semantic resources deployed by the
participants.

\secref{sec:results} suggests a few ways of interpreting and analyzing
the results. Long-term lessons will undoubtedly emerge from the workshop
discussion. One optimistic-pessimistic conclusion concerns the size
of the training data. The notable gain TD3 $\rightarrow$ TD4 suggests that
even more data would be even better, but that is so much easier said
than done: it took the organizers in excess of 1000 person-hours to
pin down the problem, hone the guidelines and relation definitions,
construct sufficient amounts of trustworthy training data, and run
the task\ldots

\bibliographystyle{acl}
\bibliography{semeval2multiway_short}

\end{document}